# Automorphism Groups of Graphical Models and Lifted Variational Inference

**Hung Hai Bui**
Natural Language Understanding Lab
Nuance Communications
bui.h.hung@gmail.com

**Tuyen N. Huynh**
Artificial Intelligence Center
SRI International
huynh@ai.sri.com

**Sebastian Riedel**
Department of Computer Science
University College London
sebastian.riedel@gmail.com

## Abstract

Using the theory of group action, we first introduce the concept of the *automorphism group* of an exponential family or a graphical model, thus formalizing the general notion of symmetry of a probabilistic model. This automorphism group provides a precise mathematical framework for lifted inference in the general exponential family. Its group action partitions the set of random variables and feature functions into equivalent classes (called orbits) having identical marginals and expectations. Then the inference problem is effectively reduced to that of computing marginals or expectations for each class, thus avoiding the need to deal with each individual variable or feature. We demonstrate the usefulness of this general framework in lifting two classes of variational approximation for maximum a posteriori (MAP) inference: local linear programming (LP) relaxation and local LP relaxation with cycle constraints; the latter yields the first lifted variational inference algorithm that operates on a bound tighter than the local constraints.

## 1  Introduction

Classical approaches to probabilistic inference—an area now reasonably well understood—have traditionally exploited low tree-width and sparsity of the graphical model for efficient exact and approximate inference. A more recent approach known as *lifted inference* [4, 16, 7, 8] has demonstrated the possibility to perform very efficient inference in highly-connected, but *symmetric* models, such as those arising in the context of relational (first-order) probabilistic models.

Symmetry is the essential element of lifted inference. But currently, no formally defined notion of symmetry of a probabilistic model exists, and thus no formal account of what "exploiting symmetry" means in lifted inference has been defined. As a result, most previous work has derived lifted versions of existing propositional algorithms from a *procedural* perspective: for models that exhibit symmetries, propositional inference algorithms tend to perform the same computations several times, and their lifted counterparts are designed to perform these operation once. This approach severely limits the theoretical understanding of the nature of lifted inference. In practice, this approach also limits the class of inference algorithms that we can lift. For example, many ground inference updates (e.g., asynchronous belief propagation, max-product linear programming (MPLP) [5]) are made in a sequence that breaks the symmetry of the original model. Likewise, with the advance in modern optimization, many algorithms rely on off-the-shelf solvers in their inner loop, and lifting these solvers is not practical.

In this work, we propose an alternative approach: rather than lifting inference *algorithms*, we lift their *variational formulations*, the optimization problems that variational inference algorithms seek to solve. These lifted formulations can then be tackled with the usual optimization toolbox (off-the-shelf solvers, cutting plane algorithms, dual block coordinate descent updates etc.). If the original model exhibits symmetry, then the lifted formulations will generally be more compact than their propositional counterparts, and hence their optimization is likely to be more efficient. This *declarative* approach to lifting gives rise to a new class of algorithms, including the first lifted variational algorithm that operates on a bound tighter than the local constraints.

This paper is divided into three parts: In the first part, we show how to find a *lifting partition*: sets of random variables and feature functions that have identical expectations. We present a formal account of symmetry in graphical models through automorphism groups of exponential families. When there is parameter-tying, the automorphism group leads to a subgroup, termed the *lifting group*, which also captures symmetry in the parameters. By linking the lifting group to the well-known subject of *graph automorphisms* [10, 6], we can leverage off-the-shelf tools to find lifting partitions as orbits of the lifting group. Further, by connecting the lifting group to *renaming* permutations of logical constants in Markov Logic Network (MLN) [14], we find lifting partitions without unrolling the MLN. In work done concurrently and independently from ours,

Niepert [12, 13] presented similar ideas for exploiting orbits of permutation groups in lifting Markov Chain Monte Carlo (MCMC) algorithms. Though the ideas are similar, unique to our contribution is the rigorously defined automorphism group of a general exponential family that enables formal proofs of all subsequent results.

In the second part, we are given a lifting partition, and we use it to collapse the variational variables and constraint set. In particular, we investigate two popular variational relaxations of MAP inference. The first one is based on the local polytope, and the second one is based on a tightening of the local polytope with cycle constraints. For the latter, we also develop a lifted separation oracle to find violated constraints in the reduced yet still exponential lifted cycle polytope.

In the third part, we evaluate the novel algorithms that our framework gives rise to. Using an off-the-shelf LP solver, we show that for models with symmetry, lifted MAP in the local polytope is more efficient than propositional MAP. Likewise, for models with symmetry and repulsion, the lifted cycle polytope yields more accurate results than its local counterpart, and requires less runtime than the propositional version. Finally, we show the effectiveness of the renaming approach to finding lifting partitions. Although the proofs are non-trivial, due to space restrictions, they are omitted but can be found in [3].

## 2 Background on Groups and Graph Automorphisms

A *partition* $\Delta = \{\Delta_1 \ldots \Delta_k\}$ of a set $V$ is a set of disjoint nonempty subsets of $V$ whose union is $V$. Each element $\Delta_i$ is called a *cell*; $|\Delta|$ is thus the number of cells or the *size* of the partition. A partition $\Delta$ defines an equivalence relation $\stackrel{\Delta}{\sim}$ on $V$ by letting $u \stackrel{\Delta}{\sim} v$ iff $u$ and $v$ are in the same cell. A partition $\Lambda$ is finer than $\Delta$ if every cell of $\Lambda$ is a subset of some cell of $\Delta$.

We now briefly review the important concepts in group theory and graph automorphisms [6]. A mathematical *group* $(\mathbb{G}, \cdot)$ is a non-empty set $\mathbb{G}$ containing an identity element, denoted by $\mathbf{1}$, and a binary operation $\cdot$ which is associative and closed in $\mathbb{G}$. The group identity satisfies $\forall g \in \mathbb{G}, \mathbf{1} \cdot g = g \cdot \mathbf{1} = g$, and every element of $\mathbb{G}$ is invertible, i.e., $\exists g^{-1}$ such that $g \cdot g^{-1} = g^{-1} \cdot g = \mathbf{1}$. A group containing $\mathbf{1}$ as its only element is called a trivial *group*. A *subgroup* of $\mathbb{G}$ is a subset of $\mathbb{G}$ that forms a group with the same binary operation as $\mathbb{G}$. We write $\mathbb{G}_1 \leq \mathbb{G}_2$ when $\mathbb{G}_1$ is a subgroup[1] of $\mathbb{G}_2$.

A permutation of a set $V$ is a bijective mapping from $V$ to itself. Two permutations can be composed together via the usual composition of two mappings. Any set of permutations (on $V$) that contains the identity permutation and is closed under composition and taking inverse thus forms a group. The set of *all* permutations of $V$ is called the *symmetric group* $\mathbb{S}(V)$. The symmetric group $\mathbb{S}_n$ is the set of all permutations of $\{1, 2, \ldots, n\}$. For a permutation $\pi \in \mathbb{S}_n$, $\pi(i)$ is the image of $i$ under $\pi$. For each vector $x \in \mathcal{X}^n$, the vector $x$ permuted by $\pi$, denoted by $x^\pi$, is $(x_{\pi(1)} \ldots x_{\pi(n)})$; for a set $A \subset \mathcal{X}^n$, the set $A$ permuted by $\pi$, denoted by $A^\pi$ is $\{x^\pi | x \in A\}$.

A subgroup $\mathbb{G}$ of $\mathbb{S}(V)$ induces the following equivalence relation on $V$: $v \sim v'$ iff there exists $g \in \mathbb{G}$ such that $g(v) = v'$ (the fact that $\sim$ is an equivalence relation follows from the definition of a group). $\mathbb{G}$ therefore induces a partition on $V$, called the *orbit partition,* denoted by $\text{Orb}_\mathbb{G}(V)$. The *orbit* of an element $v \in V$ is the set of elements in $V$ equivalent to $v$: $\text{orb}_\mathbb{G}(v) = \{v' \in \mathcal{V} | v' \sim v\}$.

A group $\mathbb{G}$ can induce an orbit partition on any set $U$ as long as members of $\mathbb{G}$ can be viewed as (not necessarily distinct) permutations of $U$. In this case, there is a group homomorphism from $\mathbb{G}$ to a subgroup of $\mathbb{S}(U)$, and the group $\mathbb{G}$ is said to *act* on the set $U$. A subgroup $\mathbb{G}_1 \leq \mathbb{G}$ will also act on $U$ and induces a finer orbit partition. Given a set element $u \in U$ and a group element $g \in \mathbb{G}$, if $g(u) = u$ then $g$ is said to stabilize $u$. If $\forall g \in \mathbb{G}, g(u) = u$, then the group $\mathbb{G}$ is said to stabilize $u$.

Group action is a powerful concept since it allows the same group $\mathbb{G}$ to act (hence induce orbit partitions) on many different sets. For example, $\mathbb{S}_n$ acts on the set of $n$-dimension vectors $\mathcal{X}^n$ via the action $\pi(x) = x^\pi$. $\mathbb{S}_n$ also acts on the set of $n$-vertex graphs in the following way. Every permutation $\pi \in \mathbb{S}_n$ transforms a graph $\mathfrak{G}$ to its isomorphic variant $\mathfrak{G}'$ (i.e., $\{i, j\}$ is an edge in $\mathfrak{G}$ iff $\{\pi(i), \pi(j)\}$ is an edge in $\mathfrak{G}'$). Hence, it can be viewed as a bijection (permutation) on the set of $n$-vertex graphs. If $\pi(\mathfrak{G}) = \mathfrak{G}$ then $\pi$ stabilizes $\mathfrak{G}$ and is called an *automorphism* of the graph $\mathfrak{G}$. The set of all automorphisms of $\mathfrak{G}$ forms a group named the *automorphism group* of $\mathfrak{G}$, denoted by $\mathbb{A}(\mathfrak{G})$ (see Figure 1). It is clear that $\mathbb{A}(\mathfrak{G})$ is a subgroup of $\mathbb{S}_n$. The cardinality of $\mathbb{A}(\mathfrak{G})$ indicates the level of symmetry in $\mathfrak{G}$. If $\mathbb{A}(\mathfrak{G})$ is the trivial group then $\mathfrak{G}$ is asymmetric; if $\mathbb{A}(\mathfrak{G}) = \mathbb{S}_n$ then $\mathfrak{G}$ either is fully connected or has no edges. This concept of graph automorphism directly generalizes to graphs with additional structures such as directions, colors, etc.

If we now ask what elements of $\mathfrak{G}$ are indistinguishable up to symmetry, the automorphism group $\mathbb{A}(\mathfrak{G})$ can give us the precise answer. For example, if $v'$ can be obtained from a node $v$ via some permutation $\pi$ in $\mathbb{A}(\mathfrak{G})$, then these two nodes are indistinguishable and must have the same the graph properties (e.g., degree, averaged distance to other nodes, etc.). $\mathbb{A}(\mathfrak{G})$ thus partitions the set of nodes $V$ into the node-orbits $\text{Orb}_{\mathbb{A}(\mathfrak{G})}(V)$ where each node orbit is a set of vertices equivalent to one another up to some node relabeling. Furthermore, $\mathbb{A}(\mathfrak{G})$ also acts on the set of graph edges $E$ of $\mathfrak{G}$ by letting $\pi(\{u, v\}) = \{\pi(u), \pi(v)\}$ and this action partitions $E$ into a set of edge-orbits $\text{Orb}_{\mathbb{A}(\mathfrak{G})}(E)$. Similarly, we can also obtain the set of arc-orbits $\text{Orb}_{\mathbb{A}(\mathfrak{G})}(\overrightarrow{E})$.

Computing the automorphism group of a graph is as difficult as determining whether two graphs are isomorphic, a problem that is known to be in NP, but for which it is unknown whether it has a polynomial time algorithm or is NP-complete. In practice, efficient computer programs, such as *nauty*[2] [10], exist for computing automorphism groups of graphs.

---

[1] We use the notation $\mathbb{G}_1 \preceq \mathbb{G}_2$ to mean $\mathbb{G}_1$ is isomorphic to a subgroup of $\mathbb{G}_2$.

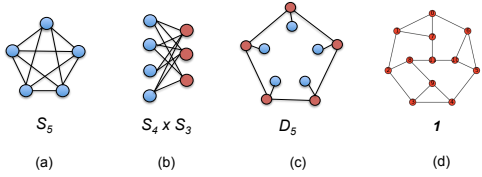

Figure 1: Graphs and their automorphism groups: (a) $\mathbb{A}(K_5) = \mathbb{S}_5$; (b) $\mathbb{A}(K_{4\times 3}) = \mathbb{S}_4 \times \mathbb{S}_3$; (c) this graph can be rotated or flipped, yielding the automorphism dihedral group $D_5$; and (d) this is known as the Frucht's graph, a regular but asymmetric graph. Blue and red colors in (a)-(c) denote different node orbits.

## 3 Symmetry of the Exponential Family

### 3.1 Exponential Family and Graphical Model

Consider an exponential family over $n$ random variables $(x_i)_{i \in \mathcal{V}}$ where $\mathcal{V} = \{1 \ldots n\}$, $x_i \in \mathcal{X}$ with density function

$$\mathcal{F}(x \,|\, \theta) = h(x) \exp\left(\langle \Phi(x), \theta \rangle - A(\theta)\right)$$

where $h$ is the base density, $\Phi(x) = (\phi_j(x))_{j \in \mathcal{I}}$, $\mathcal{I} = \{1, 2, \ldots, m\}$ is an $m$-dimensional feature vector, $\theta \in \mathbb{R}^m$ is the natural parameter, and $A(\theta)$ the log-partition function. Let $\Theta = \{\theta \,|\, A(\theta) < \infty\}$ be the set of natural parameters, $\mathcal{M} = \{\mu \in \mathbb{R}^m \,|\, \exists p, \ \mu = \mathbb{E}_p \Phi(x)\}$ the set of realizable mean parameters, $A^* : \mathcal{M} \to \mathbb{R}$ the convex dual of $A$, and $\mathbf{m} : \Theta \to \mathcal{M}$ the mean parameter mapping that maps $\theta \mapsto \mathbf{m}(\theta) = \mathbb{E}_\theta \Phi(x)$. Note that $\mathbf{m}(\Theta) = \text{ri}\,\mathcal{M}$ is the relative interior of $\mathcal{M}$. For more details, see [19].

Often, a feature function $\phi_i$ depends only on a subset of the variables in $\mathcal{V}$. In this case we will write $\phi_i$ more compactly in factorized form as $\phi_i(x) = f_i(x_{i_1} \ldots x_{i_K})$ where the indices $i_j$ are distinct, $i_1 < i_2 \ldots < i_K$, and $f_i$ cannot be reduced further, i.e., it must depend on all of its arguments. To keep track of variable indices of arguments of $f_i$, we let $scope(f_i)$ denote its set of arguments, $\eta_i(k) = i_k$ the $k$-th argument and $|\eta_i|$ its number of arguments. Factored forms of features can be encoded as a hypergraph $\mathcal{G}\,[\mathcal{F}]$ of $\mathcal{F}$ (called the graph structure or graphical model of $\mathcal{F}$) with nodes $\mathcal{V}$, and hyperedges (clusters) $\{C \,|\, \exists i, scope(f_i) = C\}$. For models with pairwise features, $\mathcal{G}$ is a standard graph.

For discrete random variables (i.e., $\mathcal{X}$ is finite), we often want to work with the overcomplete family $\mathcal{F}^o$ that we now describe for the case with pairwise features. The set of overcomplete features $\mathcal{I}^o$ are indicator functions on the nodes and edges of the graphical model $\mathcal{G}$ of $\mathcal{F}$: $\phi^o_{u:t}(x) = \mathbb{I}\{x_u = t\}, t \in \mathcal{X}$ for each node $u \in V(\mathcal{G})$; and $\phi^o_{\{u:t,v:t'\}}(x) = \mathbb{I}\{x_u = t, x_v = t'\}, t, t' \in \mathcal{X}$ for each edge $\{u,v\} \in E(\mathcal{G})$. The set of overcomplete realizable mean parameters $\mathcal{M}^o$ is also called the *marginal polytope* because the overcomplete mean param-eter corresponds to node and edge marginal probabilities. Given a parameter $\theta$, the transformation of $\mathcal{F}(x|\theta)$ to its overcomplete representation is done by letting $\theta^o$ be the corresponding parameter in the overcomplete family: $\theta^o_{u:t} = \sum_{i\, \text{s.t.}\, scope(f_i) = \{u\}} f_i(t) \theta_i$ and (assuming $u < v$) $\theta^o_{\{u:t,v:t'\}} = \sum_{i\, \text{s.t.}\, scope(f_i) = \{u,v\}} f_i(t, t') \theta_i$. Verifying that $\mathcal{F}^o(x|\theta^o) = \mathcal{F}(x|\theta)$ is straightforward.

[2]*http://cs.anu.edu.au/people/bdm/nauty/*

### 3.2 Automorphism Group of an Exponential Family

We define the symmetry of an exponential family $\mathcal{F}$ as the group of transformations that preserve $\mathcal{F}$ (hence preserve $h$ and $\Phi$). The kind of transformation used will be a pair of permutations $(\pi, \gamma)$ where $\pi$ permutes the set of variables and $\gamma$ permutes the feature vector.

**Definition 1.** An automorphism of the exponential family $F$ is a pair of permutations $(\pi, \gamma)$ where $\pi \in \mathbb{S}_n$, $\gamma \in \mathbb{S}_m$ such that for all vectors $x$: $h(x^\pi) = h(x)$ and $\Phi^{\gamma^{-1}}(x^\pi) = \Phi(x)$ (or equivalently, $\Phi(x^\pi) = \Phi^\gamma(x)$).

Showing that the set of all automorphisms of $\mathcal{F}$, denoted by $\mathbb{A}[\mathcal{F}]$, forms a subgroup of $\mathbb{S}_n \times \mathbb{S}_m$ is straightforward. This group acts on $\mathcal{I}$ by the permuting action of $\gamma$, and on $\mathcal{V}$ by the permuting action of $\pi$. In the remainder of this paper, $h$ is always a symmetric function (e.g., $h \equiv 1$); therefore, the condition $h(x^\pi) = h(x)$ automatically holds.

**Example 1.** Let $\mathcal{V} = \{1 \ldots 4\}$ and $\Phi = \{f_1 \ldots f_5\}$ where $f_1(x_1, x_2) = x_1(1-x_2)$, $f_2(x_1, x_3) = x_1(1-x_3)$, $f_3(x_2, x_3) = x_2 x_3$, $f_4(x_2, x_4) = x_4(1-x_2)$, $f_5(x_3, x_4) = x_4(1-x_3)$. Then $\pi = (1 \leftrightarrow 4)(2 \leftrightarrow 3)$, $\gamma = (1 \leftrightarrow 5)(2 \leftrightarrow 4)$ form an automorphism of $\mathcal{F}$, since $\Phi^{\gamma^{-1}}(x^\pi) = (\phi_5(x_4 \ldots x_1), \phi_4(x_4 \ldots x_1), \ldots, \phi_1(x_4 \ldots x_1)) = (f_5(x_2, x_1), f_4(x_3, x_1), f_3(x_3, x_2), f_2(x_4, x_2), f_1(x_4, x_3)) = (x_1(1-x_2), x_1(1-x_3), x_3 x_2, x_4(1-x_2), x_4(1-x_3)) = \Phi(x_1 \ldots x_4)$.

An automorphism as defined above preserves a number of key characteristics of the exponential family $\mathcal{F}$ (such as its natural parameter space, its mean parameter space, and its log-partition function), as shown in the following theorem.

**Theorem 1.** *If $(\pi, \gamma) \in \mathbb{A}[\mathcal{F}]$ then*

1. $\pi \in \mathbb{A}(\mathcal{G}[\mathcal{F}])$, i.e. $\pi$ is an automorphism of the graphical model graph $\mathcal{G}[\mathcal{F}]$.

2. $\Theta^\gamma = \Theta$ and $A(\theta^\gamma) = A(\theta)$ for all $\theta \in \Theta$.

3. $\mathcal{F}(x^\pi | \theta^\gamma) = \mathcal{F}(x|\theta)$ for all $x \in \mathcal{X}^n$, $\theta \in \Theta$.

4. $\mathbf{m}^\gamma(\theta) = \mathbf{m}(\theta^\gamma)$ for all $\theta \in \Theta$.

5. $\mathcal{M}^\gamma = \mathcal{M}$ and $A^*(\mu^\gamma) = A^*(\mu)$ for all $\mu \in \mathcal{M}$.

### 3.3 Parameter Tying and the Lifting Group

We now consider a parameter-tying setting where some components of $\theta$ are the same. Formally, a partition $\Delta$ of $\mathcal{I}$ is called the *parameter-tying partition* iff $j \stackrel{\Delta}{\sim} j' \Rightarrow \theta_j = \theta_{j'}$. Let $\mathbb{R}^m_\Delta$ denote the subspace $\left\{r \in \mathbb{R}^m \,|\, r_j = r_{j'} \text{ if } j \stackrel{\Delta}{\sim} j'\right\}$. For any set $S \subset \mathbb{R}^m$, let

$S_\Delta$ denote the set intersection $S \cap \mathbb{R}^m_\Delta$. Parameter tying is equivalent to restricting the natural parameter $\theta$ to the set $\Theta_\Delta$. This is also equivalent to working with a different exponential family with $|\Delta|$ aggregating features $\left(\sum_{j \in \Delta_i} \phi_j\right)_i$. While this family has fewer parameters, it is not obvious how it would help inference; moreover, in working directly with the aggregation features, the structure of the original family is lost. Our goal is to study how parameter-tying, coupled with the symmetry of the family $\mathcal{F}$, can lead to more efficient inference.

The automorphism group $\mathbb{A}[\mathcal{F}]$ preserves the family of distributions $\mathcal{F}$; however, this group does not take any specific parameter $\theta$ into account. Of special interest is the set of automorphisms that also preserve $\theta$ for every tied parameter $\theta \in \Theta_\Delta$. We will now formalize this concept. Given a partition $\Delta$, a permutation $\lambda$ on $\mathcal{I}$ is consistent with $\Delta$ iff $\lambda$ permutes only among elements of the same cell of $\Delta$. Clearly, for all $\theta \in \Theta_\Delta$, $\theta^\lambda = \theta$. If $\mathbb{G}$ is a group acting on $\mathcal{I}$, we let $\mathbb{G}_\Delta$ denote the set of group elements whose actions are consistent with $\Delta$, that is $\mathbb{G}_\Delta = \left\{g \in \mathbb{G} | \forall u \in \mathcal{I}, g(u) \stackrel{\Delta}{\sim} u\right\}$. It is straightforward to verify that $\mathbb{G}_\Delta$ is a subgroup of $\mathbb{G}$.

**Definition 2.** (Lifting Group) The lifting group corresponding to the parameter-tying partition $\Delta$ is $\mathbb{A}_\Delta(\mathcal{F})$, the subgroup of $\mathbb{A}[\mathcal{F}]$ whose member's action is consistent with $\Delta$.

The lifting group $\mathbb{A}_\Delta(\mathcal{F})$ thus stabilizes not just the family $\mathcal{F}$, but also every parameter $\theta \in \Theta_\Delta$. Furthermore, features in the same orbit induced by the lifting group must have the same expectation (a consequence of theorem 1, part 4). As we shall see in the later section, the lifting group $\mathbb{A}_\Delta(\mathcal{F})$ and its induced orbit partitions on the set of variables and features play a central role in our lifted variational inference framework.

## 4 Detecting Symmetries in Exponential Families

We now discuss the computation of the lifting group $\mathbb{A}_\Delta(\mathcal{F})$ and its orbit partitions. In practice, computing and working with a subgroup of the lifting group suffice.

### 4.1 Detecting Symmetries via Graph Automorphisms

Our first approach is to construct a suitable graph whose automorphism group is guaranteed to be a subgroup of $\mathbb{A}_\Delta(\mathcal{F})$, and thus any tool and algorithm for computing graph automorphism can be applied. The constructed graph resembles a factor graph representation of $\mathcal{F}$. However, we also use colors of factor nodes to mark feature functions that are both identical and in the same cell of $\Delta$, and colors of edges to encode symmetry of the feature functions themselves.

**Definition 3.** The colored factor graph induced by $\mathcal{F}$ and $\Delta$, denoted by $\mathfrak{G}_\Delta[\mathcal{F}]$ is a bipartite graph with nodes $V(\mathfrak{G}) = \{x_1 \ldots x_n\} \cup \{f_1 \ldots f_m\}$ and edges $E(\mathfrak{G}) = \{\{x_{\eta_i(k)}, f_i\} \mid i \in \mathcal{I}, k = 1 \ldots |\eta_i|\}$. Variable nodes are assigned the same color which is different from the colors of factor nodes. Factor nodes $f_i$ and $f_j$ have the same color iff $f_i \equiv f_j$ and $i \stackrel{\Delta}{\sim} j$. If the function $f_i$ is symmetric, then all edges adjacent to $f_i$ have the same color; otherwise, they are colored according to the argument number of $f_i$, i.e., $\{x_{\eta_i(k)}, f_i\}$ is assigned the $k$-th color.

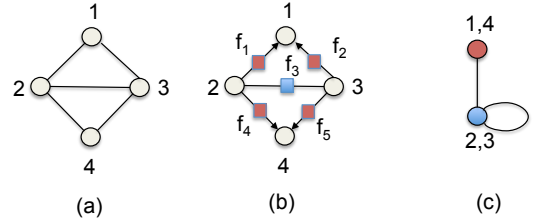

Figure 2: Graph construction for computing the lifting group and its orbits: (a) original graphical model of example 1; (b) constructed colored factor graphs, assuming all parameters are the same (arrows represent first arguments of the asymmetric factors); and (c) lifted graphical model with nodes representing node orbits and edges representing edge orbits of the original graphical model.

Figure 2 shows the construction of the colored factor graph for the exponential family in example 1 where we have assumed that all the parameters are the same.

**Theorem 2.** *The automorphism group $\mathbb{A}[\mathfrak{G}_\Delta]$ of $\mathfrak{G}_\Delta[\mathcal{F}]$ is a subgroup of $\mathbb{A}_\Delta(\mathcal{F})$, i.e., $\mathbb{A}[\mathfrak{G}_\Delta] \leq \mathbb{A}_\Delta[\mathcal{F}]$.*

Finding the automorphism group $\mathbb{A}[\mathfrak{G}_\Delta]$ of the graph $\mathfrak{G}_\Delta[\mathcal{F}]$ therefore yields a procedure to compute a subgroup of $\mathbb{A}_\Delta[\mathcal{F}]$. *Nauty*, for example, directly implements operations of computing the automorphism group of a graph and extracting the induced node orbits and edge orbits.

### 4.2 Symmetries of Markov Logic Networks

Markov Logic Network (MLN) [14] is a first-order probabilistic model that defines an exponential family on random structures (i.e., random graphs, hypergraphs, or more generally random Herbrand models of the first-order language). In this case, a subgroup of the lifting group can be obtained via the symmetry of the unobserved constants in the domain without the need to consider the ground graphical model.

An MLN is prescribed by a list of weighted formulas $F_1 \ldots F_K$ (consisting of a set of predicates, logical variables, constants, and a weight vector $\mathbf{w}$) and a logical domain $\mathcal{D} = \{a_1 \ldots a_{|\mathcal{D}|}\}$. Let $\mathcal{D}_0$ be the set of objects appearing as constants in these formulas, then $\mathcal{D}_* = \mathcal{D} \backslash \mathcal{D}_0$ is the set of objects in $\mathcal{D}$ that do not appear in these formulas. Let Gr be the set of all ground predicates $p(a_1 \ldots a_\ell)$'s. Given a substitution $s$, $F_i[s]$ denotes the result of applying the substitution $s$ to $F_i$ and is a grounding of $F_i$ if it does not contain any free logical variables. The set of all groundings of $F_i$ is $\mathrm{GrF}_i$, and let $\mathrm{GrF} = \mathrm{GrF}_1 \cup \ldots \cup \mathrm{GrF}_K$. Let $\omega$ be a truth assignment to all the ground predicates in Gr and $\mathrm{w}_i$ be the weight of the formula $F_i$. The MLN corresponds to an exponential family $\mathcal{F}_{MLN}$ where Gr is the variable index set and each grounding $F_i[s] \in \mathrm{GrF}_i$ is a feature function $\phi_{F_i[s]}(\omega) = \mathbb{I}(\omega \vDash F_i[s])$ with the associated parameter $\theta_{F_i[s]} = \mathrm{w}_i$. Since all the ground features of the formula $F_i$ have the same parameter $\mathrm{w}_i$, the MLN also induces the parameter-tying partition $\Delta_{MLN} = \{\{\phi_{F_1[s]}(\omega)\} \ldots \{\phi_{F_K[s]}(\omega)\}\}$.

Let a renaming permutation $r$ be a permutation over $\mathcal{D}$ that fixes every object in $\mathcal{D}_0$ (i.e., $r$ only permutes objects in $\mathcal{D}_*$). Thus, the set of all such renaming permutations is a group $\mathbb{G}^{re}$ isomorphic to the symmetric group $\mathbb{S}(\mathcal{D}_*)$. Consider the following action of $\mathbb{G}^{re}$ on Gr : $\pi_r$ : $p(a_1 \ldots a_\ell) \mapsto p(r(a_1) \ldots r(a_\ell))$, and the action on GrF $\gamma_r : F_i[s] \mapsto F_i[r(s)]$ where $r(s = (x_1/a_1, ..., x_k/a_k)) = (x_1/r(a_1), ..., x_k/r(a_k))$. Intuitively, $\pi_r$ and $\gamma_r$ rename the constants in each ground predicate $p(a_1 \ldots a_\ell)$ and ground formula $F_i[s]$ according to the renaming permutation $r$. The following is a consequence of Lemma 1 from Bui et al. [2].

**Theorem 3.** *For every renaming permutation $r$, $(\pi_r, \gamma_r) \in \mathbb{A}[\mathcal{F}_{MLN}]$. Further, the renaming group $\mathbb{G}^{re}$ is isomorphic to a subgroup of the MLN's lifting group: $\mathbb{G}^{re} \preceq \mathbb{A}_{\Delta_{MLN}}[\mathcal{F}_{MLN}]$.*

Orbit partitions induced by $\mathbb{G}^{re}$ on the set of predicate groundings can be derived directly from the first-order representation of an MLN without considering its ground graphical model. The size of this orbit partition depends only on the number of observed constants $|\mathcal{D}_o|$, and does not depend on actual domain size $|\mathcal{D}|$. For example, if $q(.,.)$ is a 2-ary predicate and there is one observed constant $a$, then we obtain the following partition of the groundings of $q$: $\{q(a,a)\}$, $\{q(x,x)|x \neq a\}$, $\{q(a,x)|x \neq a\}$, $\{q(x,a)|x \neq a\}$, $\{q(x,y)|x \neq y, x \neq a, y \neq a\}$. Similar partitions on the set of factors and variable clusters can also be obtained with complexity polynomial in $|\mathcal{D}_o|$ and independent of $|\mathcal{D}|$.

## 5 Lifted Variational Inference Framework

We now discuss the principle of how to exploit the symmetry of the exponential family graphical model for lifted variational inference. In the general variational inference framework [19], marginal inference is viewed as a means to compute the mean parameter $\mu = \mathbf{m}(\theta)$ given a natural parameter $\theta$ by solving the optimization problem

$$\sup_{\mu \in \mathcal{M}} \langle \theta, \mu \rangle - A^*(\mu). \tag{1}$$

For discrete models, the variational problem is more conveniently posed using the overcomplete parameterization, for marginal and MAP inference

$$\sup_{\mu^o \in \mathcal{M}^o} \langle \mu^o, \theta^o \rangle - A^{o*}(\mu^o) \tag{2}$$

$$\max_{x \in \mathcal{X}^n} \ln \mathcal{F}(x|\theta) = \sup_{\mu^o \in \mathcal{M}^o} \langle \mu^o, \theta^o \rangle + \text{const.} \tag{3}$$

We first focus on lifting the main variational problem in (1) and leave discussions of the other problems to subsection 5.3.

### 5.1 Lifting Partition

Consider the parameter-tying scenario where $\theta \in \Theta_\Delta$ for a given partition $\Delta$ on the feature set $\mathcal{I}$. With this restriction, the mean parameter by definition must lie inside $\mathbf{m}(\Theta_\Delta)$, so in theory, the domain of the variational optimization problems can be restricted to $\mathbf{m}(\Theta_\Delta)$. The main difficulty here lies in how to characterize $\mathbf{m}(\Theta_\Delta)$.

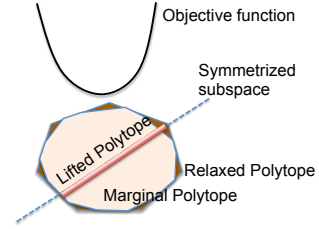

Figure 3: (Best viewed in color) Symmetrized subspace

We first make a rather intuitive observation: for general convex optimization problems with symmetric objective functions and constraints, the optimal solutions are trapped in a lower-dimensional symmetrized subspace (see Figure 5.1). This is formalized in lemma 1, whose proof makes use of the orbit-stabilizer theorem, an elementary result in group theory.

**Definition 4.** *(Lifting partition) Consider the convex optimization $\inf_{\mathbf{x} \in \mathcal{S}} J(\mathbf{x})$ where $\mathcal{S} \subset \mathbb{R}^m$ is a convex set and $J$ is a convex function. A partition $\varphi$ of $\{1 \ldots m\}$ is a lifting partition for the aforementioned problem iff $\inf_{x \in S} J(x) = \inf_{x \in S_\varphi} J(x)$ (i.e., the constraint set $S$ can be restricted to $S_\varphi = S \cap \mathbb{R}^m_\varphi$).*

**Lemma 1.** *Let $\mathbb{G}$ act on $I = \{1 \ldots m\}$, so that every $g \in \mathbb{G}$ corresponds to some permutation on $\{1 \ldots m\}$. If $S^g = S$ and $J(x^g) = J(x)$ for every $g \in \mathbb{G}$ (i.e., $\mathbb{G}$ stabilizes both $S$ and $J$) then the induced orbit partition $\mathrm{Orb}_{\mathbb{G}}(I)$ is a lifting partition for $\inf_{x \in \mathcal{S}} J(x)$.*

The second key observation is that all the above variational problems inherit the same symmetries of the parameter-tying exponential family, as captured in the lifting group $\mathbb{A}_\Delta[\mathcal{F}]$. Therefore, the lifting group will play the role of $\mathbb{G}$ in lemma 1 in lifting all of our variational problems.

Returning to (1), our general principle of lifted variational inference is captured in the following therem.

**Theorem 4.** *Let $\varphi = \varphi(\Delta) = \mathrm{Orb}_{\mathbb{A}_\Delta[\mathcal{F}]}(\mathcal{I})$. Then for all $\theta \in \Theta_\Delta$, $\varphi$ is a lifting partition for (1), i.e.*

$$\sup_{\mu \in \mathcal{M}} \langle \theta, \mu \rangle - A^*(\mu) = \sup_{\mu \in \mathcal{M}_\varphi} \langle \theta, \mu \rangle - A^*(\mu) \tag{4}$$

*Sktech of proof.* From theorem 1, $\mathbb{A}[\mathcal{F}]$ stabilizes $\mathcal{M}$ and $A^*$; further, its subgroup $\mathbb{A}_\Delta(\mathcal{F})$ stabilizes every parameter $\theta \in \Theta_\Delta$. Thus, the lifting group $\mathbb{A}_\Delta(\mathcal{F})$ stabilizes both the constraint set and the objective function of (1). Invoking lemma 1, the induced orbit partition on $\mathcal{I}$ therefore yields a lifting partition.

In (4), we call the LHS the *ground* formulation of the variational problem, and the RHS the *lifted* formulation. Let $\ell = |\varphi|$ be the number of cells of $\varphi$, the *lifted* mean parameter space $\mathcal{M}_\varphi$ then effectively lies inside an $\ell$-dimensional subspace where $\ell \leq m$. This forms the core of our principle of lifted variational inference: to perform optimization over the lower dimensional (and hopefully easier) constraint set $\mathcal{M}_\varphi$ instead of $\mathcal{M}$.

*Remark.* Because (1) has a unique solution $\mu = \mathbf{m}(\theta)$, theorem 4 implies that $\mathbf{m}(\Theta_\Delta) \subset \mathcal{M}_\varphi$. Further, the theorem also holds if we replace $\mathbb{A}_\Delta(\mathcal{F})$ with one of its subgroups $\mathbb{G}$: since $\varphi_\mathbb{G} = \mathrm{Orb}_\mathbb{G}(\mathcal{I})$ is finer than $\varphi$, it is obvious that $\varphi_\mathbb{G}$ is also a lifting partition. However, the smaller is the group $\mathbb{G}$, the finer is the lifting partition $\varphi_\mathbb{G}$, and the less symmetry can be exploited. In the extreme, $\mathbb{G}$ can be the trivial group, $\varphi_\mathbb{G}$ is the discrete partition putting each element of $\mathcal{I}$ in its own cell, and $\mathcal{M}_{\varphi_\mathbb{G}} = \mathcal{M}$, which corresponds to no lifting.

### 5.2 Characterization of $\mathcal{M}_\varphi$

We now give a characterization of the lifted mean parameter space $\mathcal{M}_\varphi$ in the case of discrete random variables. Note that $\mathcal{M}$ is the convex hull $\mathcal{M} = \mathrm{conv}\,\{\Phi(x)|x \in \mathcal{X}^n\}$ which is a polytope in $\mathbb{R}^m$, and $\mathbb{A}[\mathcal{F}]$ acts on the set of configurations $\mathcal{X}^n$ by the permuting action of $\pi$ which maps $x \mapsto x^\pi$ for $x \in \mathcal{X}^n$.

**Theorem 5.** *Let $\mathcal{O} = \mathrm{Orb}_{\mathbb{A}_\Delta[\mathcal{F}]}(\mathcal{X}^n)$ be the set of $\mathcal{X}$-configuration orbits. For each orbit $\mathcal{C} \in \mathcal{O}$, let $\bar{\Phi}(\mathcal{C}) = \frac{1}{|\mathcal{C}|}\sum_{x \in \mathcal{C}} \Phi(x)$ be the feature-centroid of all the configurations in $\mathcal{C}$. Then $\mathcal{M}_{\varphi(\Delta)} = \mathrm{conv}\,\{\bar{\Phi}(\mathcal{C})|\mathcal{C} \in \mathcal{O}\}$.*

Thus, the *lifted polytope* $\mathcal{M}_\varphi$ can have at most $|\mathcal{O}|$ extreme points. The number of configuration orbits $|\mathcal{O}|$ can be much smaller than the total number of configurations $|\mathcal{X}|^n$ when the model is highly symmetric. For example, for a fully connected graphical model with identical pairwise and unary potentials and $\mathcal{X} = \{0,1\}$ then every permutation $\pi \in \mathbb{S}_n$ is part of an automorphism; thus, every configuration with the same number of 1's belongs to the same orbit, and hence $|\mathcal{O}| = n+1$. In general, however, $|\mathcal{O}|$ often is still exponential in $n$. We discuss approximations of $\mathcal{M}_\varphi$ in Section 6.

A representation of the lifted polytope $\mathcal{M}_\varphi$ by a set of constraints in $\mathbb{R}^{|\varphi|}$ can be directly obtained from the constraints of the polytope $\mathcal{M}$. First, we enforce the constraint $\mu \in \mathbb{R}_\varphi^m$: for each cell $\varphi_j$ $(j = 1,\ldots,|\varphi|)$ of $\varphi$, let $\bar{\mu}_j$ be the common value of the variables $\mu_i$, $i \in \varphi_j$. Let $\rho$ be *the orbit mapping function* that maps each element $i \in \mathcal{I}$ to the corresponding cell $\rho(i) = j$ that contains $i$. Next, substituting $\mu_i$ by $\bar{\mu}_{\rho(i)}$ in the constraints of $\mathcal{M}$, we obtain a set of constraints in $\bar{\mu}$ (in vector form, we substitute $\mu$ by $D\bar{\mu}$ where $D_{ij} = 1$ if $i \in \varphi_j$ and 0 otherwise). In doing this, some constraints will become identical and thus redundant. In general, the number of non-redundant constraints can still be exponential.

### 5.3 Overcomplete Variational Problems

We now state analogous results in lifting the overcomplete variational problems (2) and (3) when $\mathcal{X}$ is finite. To simplify notation, we only present the case where features are unary or pairwise. As before, the lifting group $\mathbb{A}_\Delta[\mathcal{F}]$ will be used to induce a lifting partition. However, we need to define the action of this group on the set of overcomplete features $\mathcal{I}^o$.

For each automorphism $(\pi,\gamma) \in \mathbb{A}[\mathcal{F}]$, $\gamma$ gives us the permutation on $\mathcal{I}$. In order to obtain a permutation on $\mathcal{I}^o$, we will need to use $\pi$. By theorem 1, $\pi$ is an automorphism of the graphical model graph $\mathcal{G}$. Since overcomplete features naturally correspond to nodes and edges of $\mathcal{G}$, $\pi$ induces a natural bijection on $\mathcal{I}^o$ that maps $v{:}t \mapsto \pi(v){:}t$ and $\{u{:}t, v{:}t'\} \mapsto \{\pi(u){:}t, \pi(v){:}t'\}$. Define $\varphi^o = \varphi^o(\Delta) = \mathrm{Orb}_{\mathbb{A}_\Delta[\mathcal{F}]}(\mathcal{I}^o)$ to be the orbits of $\mathbb{A}_\Delta[\mathcal{F}]$ acting on the set of overcomplete features. Then

**Theorem 6.** *For all $\theta \in \Theta_\Delta$, $\varphi^o$ is a lifting partition for the variational problems (2) and (3).*

Thus, the optimization domain can be restricted to $\mathcal{M}_{\varphi^o}^o$ which we term the *lifted marginal polytope*. The cells of $\varphi^o$ are intimately connected to the node, edge and arc orbits of the graph $\mathcal{G}$ induced by $\mathbb{A}_\Delta[\mathcal{F}]$. We now list all the cells of $\varphi^o$ in the case where $\mathcal{X} = \{0,1\}$: each node orbit $\mathbf{v}$ corresponds to 2 cells $\{v : t|v \in \mathbf{v}\}$, $t \in \{0,1\}$; each edge orbit $\mathbf{e}$ corresponds to 2 cells $\{\{u:t, v:t\}\,|\,\{u,v\} \in \mathbf{e}\}$, $t \in \{0,1\}$; and each arc orbit $\mathbf{a}$ corresponds to the cell $\{\{u:0, v:1\}\,|(u,v) \in \mathbf{a}\}$. The orbit mapping function $\rho$ maps each element of $\mathcal{I}^o$ to its orbit as follows: $\rho(v{:}t) = \bar{v}{:}t$, $\rho(\{u{:}t, v{:}t\}) = \{\overline{u,v}\}{:}t$, $\rho(\{u{:}0, v{:}1\}) = (\overline{u,v}){:}01$ where $\bar{v}$ represents the node-orbit of $v$, $\{\overline{u,v}\}$ represents the edge-orbit of $\{u,v\}$ and $(\overline{u,v})$ represents the arc-orbit of $(u,v)$.

The total number of cells of $\varphi^o$ is $2|\bar{V}|+2|\bar{E}|+|\bar{A}|$ where $|\bar{V}|, |\bar{E}|$ and $|\bar{A}|$ are the number of node, edge and arc orbits of $\mathcal{G}$ (note that $|\bar{A}| \leq 2|\bar{E}|$). Therefore, in working with $\mathcal{M}_{\varphi^o}^o$, the big-$O$ order of the number of variables is reduced from the number of nodes and edges in $\mathcal{G}$ to the number of node and edge orbits.

For MAP inference, (3) is equivalent to the lifted problem $\sup_{\mu^o \in \mathcal{M}_{\varphi^o}^o} \langle \theta^o, \mu^o \rangle$. A single ground MAP solution $\hat{x}$ leads to an entire configuration orbit $\mathcal{C} = \mathrm{orb}_{\mathbb{A}_\Delta[\mathcal{F}]}(\hat{x})$ of MAP solutions. The feature-centroid $\bar{\mu}^o = \bar{\Phi}^o(\mathcal{C}) = \frac{1}{|\mathcal{C}|}\sum_{x \in \mathcal{C}} \Phi^o(x)$ then lies inside $\mathcal{M}_{\varphi^o}^o$ and is the corresponding lifted MAP solution. Furthermore, $\bar{\mu}_{v{:}t}^o = \frac{1}{|\mathbf{v}|}\sum_{v' \in \mathbf{v}} \phi_{v'{:}t}^o(\hat{x})$ is the fraction of the ground variables in $\hat{x}_\mathbf{v}$ assigned the value $t$, and similarly for pairwise features. Note that from the learning (parameter estimation) point of view, the lifted MAP solution is more useful than any single MAP solution alone.

## 6 Lifted Approximate MAP Inference

Approximate convex variational inference typically works with a tractable convex approximation of $\mathcal{M}$ and a tractable convex approximation of the negative entropy function $A^*$. In this paper we consider only lifted outer bounds of $\mathcal{M}^o$ (and thus restrict ourselves to the discrete case). We leave the problem of handling approximations of $A^*$ to future work. Our focus is the LP relaxation of the MAP inference problem (3) and its lifted formulation.

To find an approximate lifted solution, since any outer bound OUTER $\supset \mathcal{M}^o$ yields an outer bound OUTER$_{\varphi^o}$ of $\mathcal{M}_{\varphi^o}^o$, we can always relax the lifted problem and replace $\mathcal{M}_{\varphi^o}$ by OUTER$_{\varphi^o}$. But is the relaxed lifted problem on OUTER$_{\varphi^o}$ equivalent to the relaxed ground problem on OUTER? This depends on whether $\varphi^o$ is a lifting partition for the relaxed ground problem.

**Theorem 7.** *If the set OUTER = OUTER($\mathcal{G}$) depends only on the graphical model structure $\mathcal{G}$ of $\mathcal{F}$, then $\forall \theta \in \Theta_\Delta$, $\varphi^o$ is a lifting partition for the relaxed MAP problem*

$$\sup_{\mu^o \in \text{OUTER}} \langle \theta^o, \mu^o \rangle = \sup_{\mu^o \in \text{OUTER}_{\varphi^o}} \langle \theta^o, \mu^o \rangle$$

The most often used outer bound of $\mathcal{M}^o$ is the local marginal polytope LOCAL($\mathcal{G}$) [19], which enforces consistency for marginals on nodes and between nodes and edges of $\mathcal{G}$. [17, 18] used CYCLE($\mathcal{G}$), which is a tighter bound that also enforces consistency of edge marginals on the same cycle of $\mathcal{G}$. The Sherali-Adams hierarchy[3] [15] provides a sequence of outer bounds of $\mathcal{M}^o$, starting from LOCAL($\mathcal{G}$) and progressively tightening it to the exact marginal polytope $\mathcal{M}^o$. All of these outer bounds depend only on the structure of the graphical model $\mathcal{G}$, and thus the corresponding relaxed MAP problems admit $\varphi^o$ as a lifting partition. Note that with the exception when OUTER = LOCAL, equitable partitions [6] of $\mathcal{G}$ such as those used in [11] are not lifting partitions for the approximate variational problem in theorem 7.[4]

## 7 Lifted MAP Inference on the Local Polytope

We now focus on lifted approximate MAP inference using the local marginal polytope LOCAL. From this point on, we also restrict ourselves to models where the features are pairwise or unary, and the variables are binary ($\mathcal{X} = \{0, 1\}$).

We first aim to give an explicit characterization of the constraints of the lifted local polytope LOCAL$_{\varphi^o}$. The local polytope LOCAL($\mathcal{G}$) is defined as the set of locally consistent pseudo-marginals.

$$\left\{ \tau \geq 0 \middle| \begin{array}{ll} \tau_{v:0} + \tau_{v:1} = 1 & \forall v \in \mathcal{V}(\mathcal{G}) \\ \tau_{\{u:0,v:0\}} + \tau_{\{u:0,v:1\}} = \tau_{u:0} & \\ \tau_{\{u:0,v:0\}} + \tau_{\{v:0,u:1\}} = \tau_{v:0} & \forall \{u,v\} \in E(\mathcal{G}) \\ \tau_{\{u:1,v:1\}} + \tau_{\{u:0,v:1\}} = \tau_{v:1} & \\ \tau_{\{u:1,v:1\}} + \tau_{\{v:0,u:1\}} = \tau_{u:1} & \end{array} \right\}$$

Substituting $\tau_i$ by the corresponding $\bar{\tau}_{\rho(i)}$ where $\rho()$ is given in subsection 5.3, and by noting that constraints generated by $\{u,v\}$ in the same edge orbits are redundant, we obtain the constraints for the *lifted local polytope* LOCAL$_{\varphi^o}$ as follows.

$$\left\{ \bar{\tau} \geq 0 \middle| \begin{array}{ll} \bar{\tau}_{\mathbf{v}:0} + \bar{\tau}_{\mathbf{v}:1} = 1 & \forall \text{ node orbit } \mathbf{v} \\ \bar{\tau}_{\mathbf{e}:00} + \bar{\tau}_{(\overline{u,v}):01} = \bar{\tau}_{\bar{u}:0} & \\ \bar{\tau}_{\mathbf{e}:00} + \bar{\tau}_{(\overline{v,u}):01} = \bar{\tau}_{\bar{v}:0} & \forall \text{ edge orbit } \mathbf{e} \text{ with} \\ \bar{\tau}_{\mathbf{e}:11} + \bar{\tau}_{(\overline{u,v}):01} = \bar{\tau}_{\bar{v}:1} & \{u,v\} \text{ a representative of } \mathbf{e} \\ \bar{\tau}_{\mathbf{e}:11} + \bar{\tau}_{(\overline{v,u}):01} = \bar{\tau}_{\bar{u}:1} & \end{array} \right\}$$

---

[3]A note about terminology: Following the tradition in lifted inference, this paper uses the term *lift* to refer to the exploitation of symmetry for avoiding doing inference on the *ground* model. It is unfortunate that the term *lift* has also been used in the context of coming up with better bounds for the marginal polytopes. There, *lift* (as in lift-and-project) means to move to a higher dimensional space where constraints can be more easily expressed with auxiliary variables.

[4]As a counter example, consider a graphical model whose structure is the Frucht graph (Fig. 1(d)). Since this is a regular graph, LOCAL approximation yields identical constraints for every node. However, the nodes on this graph participate in cycles of different length, hence are subject to different cycle constraints.

Thus, the number of constraints needed to describe the lifted local polytope LOCAL$_{\varphi^o}$ is $O(|\bar{V}| + |\bar{E}|)$. Similar to the ground problem, these constraints can be derived from a graph representation of the node and edge orbits. Define the *lifted graph* $\bar{\mathcal{G}}$ to be a graph whose nodes are the set of node orbits $\bar{V}$ of $\mathcal{G}$. For each edge orbit $\mathbf{e}$ with a representative $\{u,v\} \in \mathbf{e}$, there is a corresponding edge on $\bar{\mathcal{G}}$ that connects the two node orbits $\bar{u}$ and $\bar{v}$. Note that unlike $\mathcal{G}$, the lifted graph $\bar{\mathcal{G}}$ in general is not a simple graph and can contain self-loops and multi-edges between two nodes. Figure 2(a) and (c) show the ground graphical model $\mathcal{G}$ and the lifted graph $\bar{\mathcal{G}}$ for the example 1.

Next consider the linear objective function $\langle \theta^o, \tau \rangle$. Substituting $\tau_i$ by the corresponding $\bar{\tau}_{\rho(i)}$, we can rewrite the objective function in terms of $\bar{\tau}$ as $\langle \bar{\theta}, \bar{\tau} \rangle$ where the coefficients $\bar{\theta}$ are defined on nodes and edges of the lifted graph $\bar{\mathcal{G}}$ as follows. For each node orbit $\mathbf{v}$, $\bar{\theta}_{\mathbf{v}:t} = \sum_{v' \in \mathbf{v}} \theta^o_{v':t} = |\bar{v}|\theta^o_{v:t}$ where $t \in \{0,1\}$ and $v$ is any representative member of $\mathbf{v}$. For each edge orbit $\mathbf{e}$ with a representative $\{u,v\} \in \mathbf{e}$, $\bar{\theta}_{\mathbf{e}:tt} = \sum_{\{u',v'\} \in \mathbf{e}} \theta^o_{\{u':t,v':t\}} = |\mathbf{e}|\theta^o_{\{u:t,v:t\}}$ where $t \in \{0,1\}$, $\bar{\theta}_{(\overline{u,v}):01} = \sum_{(u',v') \in (\overline{u,v})} \theta^o_{\{u':0,v':1\}} = |(\overline{u,v})|\theta^o_{\{u:0,v:1\}}$. Note that typically the two arc-orbits $(\overline{u,v})$ and $(\overline{v,u})$ are not the same, in which case $|(\overline{u,v})| = |(\overline{v,u})| = |\mathbf{e}|$. However, in the case $(\overline{u,v}) = (\overline{v,u})$, then $|(\overline{u,v})| = |(\overline{v,u})| = 2|\mathbf{e}|$.

We have shown that the lifted formulation for MAP inference on the local polytope can be described in terms of the lifted variables $\bar{\tau}$ and the lifted parameters $\bar{\theta}$. These lifted variables and parameters are associated with the orbits of the ground graphical model. Thus, the derived lifted formulation can also be read out directly from the lifted graph $\bar{\mathcal{G}}$. In fact, the derived lifted formulation is the local relaxed MAP problem of the lifted graphical model $\bar{\mathcal{G}}$. Therefore, any algorithm for solving the local relaxed MAP problem on $\mathcal{G}$ can also be used to solve the derived lifted formulation on $\bar{\mathcal{G}}$. For example, performing coordinate descent in the dual formulation [5] of the lifted local LP yields the lifted MPLP. Note that MPLP is an asynchronous message passing algorithms that cannot be lifted by grouping identical messages.

## 8 Beyond Local Polytope: Lifted MAP Inference with Cycle Inequalities

We now discuss lifting the MAP relaxation on CYCLE($\mathcal{G}$), a bound obtained by tightening LOCAL($\mathcal{G}$) with an additional set of linear constraints that hold on cycles of the graphical model structure $\mathcal{G}$, called cycle constraints [17]. These constraints mean the number of cuts (transitions from 0 to 1 or vice versa) in any configuration on a cycle of $\mathcal{G}$ must be even. Cycle constraints can be expressed as linear constraints as follows. For every cycle $C$ (set of edges that form a cycle in $\mathcal{G}$) and every odd-sized subset $F \subseteq C$

$$\sum_{\{u,v\} \in F} nocut(\{u,v\}, \tau) + \sum_{\{u,v\} \in C \setminus F} cut(\{u,v\}, \tau) \geq 1 \quad (5)$$

where $nocut(\{u,v\},\tau) = \tau_{\{u:0,v:0\}} + \tau_{\{u:1,v:1\}}$ and $cut(\{u,v\},\tau) = \tau_{\{u:0,v:1\}} + \tau_{\{v:0,u:1\}}$.

Theorem 7 guarantees that MAP inference on CYCLE can be lifted by restricting the feasible domain to CYCLE$_{\varphi^\circ}$, which we term the *lifted cycle polytope*. Substituting the original variables $\tau$ by the lifted variables $\bar\tau$, we obtain the *lifted cycle constraints* in terms of $\bar\tau$

$$\sum_{\{u,v\}\in F} nocut(\{\overline{u,v}\},\bar\tau) + \sum_{\{u,v\}\in C\setminus F} cut(\{\overline{u,v}\},\bar\tau) \geq 1 \quad (6)$$

where $nocut(\{\overline{u,v}\},\bar\tau) = \bar\tau_{\{\overline{u,v}\}:00} + \bar\tau_{\{\overline{u,v}\}:11}$ and $cut(\{\overline{u,v}\},\bar\tau) = \bar\tau_{(\overline{u,v}):01} + \bar\tau_{(\overline{v,u}):01}$ where $(\overline{u,v})$ and $(\overline{v,u})$ are the arc-orbits corresponding to the node-orbit $\{\overline{u,v}\}$.

### 8.1 Lifted Cycle Constraints on All Cycles Passing Through a Fixed Node

It is not possible to extract all lifted cycle constraints just by examining the lifted graphical model $\bar{\mathcal{G}}$ since there could be cycles in $\bar{\mathcal{G}}$ that do not correspond to any cycles in $\mathcal{G}$. However, we can characterize all constraints on all cycles passing through a fix node $i$ in $\mathcal{G}$.

Let Cyc$[i]$ be the set of (ground) cycle constraints generated from all cycles passing through $i$. A cycle is simple if it does not intersect with itself or contain repeated edges; [17] considers only simple cycles, but we will also consider any cycle, including non-simple cycles in Cyc$[i]$. Adding non-simple cycles to the mix does not change the story since constraints on non-simple cycles of $\mathcal{G}$ are redundant. We now give a precise characterization of $\overline{\text{Cyc}}[i]$, the set of lifted cycle constraints obtained by lifting all cycle constraints in Cyc$[i]$ via the transformation from (5) to (6).

The lifted graph fixing $i$, $\bar{\mathcal{G}}[i]$ is defined as follows. Let $\mathbb{A}_\Delta[\mathcal{F},i]$ be the subgroup of $\mathbb{A}_\Delta[\mathcal{F}]$ that fixes $i$, that is $\pi(i) = i$. The set of nodes of $\bar{\mathcal{G}}[i]$ is the set of node orbits $\bar V[i]$ of $\mathcal{G}$ induced by $\mathbb{A}_\Delta[\mathcal{F},i]$, and the set of edges is the set of edge orbits $\bar E[i]$ of $\mathcal{G}$. Each edge orbit connects to the orbits of the two adjacent nodes (which could form just one node orbit). Since $i$ is fixed, $\{i\}$ is a node orbit, and hence is a node on $\bar{\mathcal{G}}[i]$. Note that $\bar{\mathcal{G}}[i]$ in general is not a simple graph: it can have multi-edges and loops.

**Theorem 8.** *Let $\bar C$ be a cycle (not necessarily simple) in $\bar{\mathcal{G}}[i]$ that passes through the node $\{i\}$. For any odd-sized $\bar F \subset \bar C$*

$$\sum_{\mathbf{e}\in\bar F} nocut(\mathbf{e},\bar\tau) + \sum_{\mathbf{e}\in\bar C\setminus\bar F} cut(\mathbf{e},\bar\tau) \geq 1 \quad (7)$$

*is a constraint in $\overline{\text{Cyc}}[i]$. Further, all constraints in $\overline{\text{Cyc}}[i]$ can be expressed this way.*

### 8.2 Separation of Lifted Cycle Constraints

While the number of cycle constraints may be reduced significantly in the lifted space, it may still be computationally expensive to list all of them. To address this issue, we follow [17] and employ a cutting plane approach in which we find and add only the most violated lifted cycle constraint in each iteration (separation operation).

For finding the most violated lifted cycle constraint, we propose a lifted version of the method presented by [17], which performs the separation by iterating over the nodes of the graph $\mathcal{G}$ and for each node $i$ finds the most violated cycle constraint from all cycles passing through $i$. Theorem 8 suggests that all lifted cycle constraints in $\overline{\text{Cyc}}[i]$ can be separated by mirroring $\bar{\mathcal{G}}[i]$ and performing a shortest path search from $\{i\}$ to its mirrored node, similar to the way separation is performed on ground cycle constraints [17].

To find the most violated lifted cycle constraint, we could first find the most violated lifted cycle constraint $C_i$ in $\overline{\text{Cyc}}[i]$ for each node $i$, and then take the most violated constraints over all $C_i$. However, note that if $i$ and $i'$ are in the same node orbit, then $\overline{\text{Cyc}}[i] = \overline{\text{Cyc}}[i']$. Hence, we can perform separation using the following algorithm:

1. For each node orbit $\bar v \in \bar V$, choose a representative $i \in \bar v$ and find its most violated lifted cycle constraint $C_{\bar v} \in \overline{\text{Cyc}}[i]$ using a shortest path algorithm on the mirror graph of $\bar{\mathcal{G}}[i]$.

2. Return the most violated constraint over all $C_{\bar v}$.

Notice that both $\bar{\mathcal{G}}[i]$ and its mirror graph have to be calculated only once per graph. In each separation iteration we can reuse these structures, provided that we adapt the edge weights in the mirror graph according to the current marginals.

## 9 Experiments

First, we evaluate methods for detecting symmetries described in Section 4 on the "Friends & Smokers" MLN[5] [16]. The first method (*nauty*) grounds the MLN then finds a lifting partition. The second (*renaming*) does not require grounding, but uses the renaming group to find a lifting partition. Table 1 presents the results for varying domain sizes where for a random 10% of all people it is known whether they smoke or not. Although *nauty* finds a more compact lifted graph, it takes significantly more time than using the renaming group. For this reason, our subsequent experiment only makes use of the renaming group and orbits.[6]

Figure 4 shows the run time performance of MAP inference using local and cycle LP formulations (both ground and lifted algorithms use the off-the-shelf Gurobi LP solver). For cutting plane, we use the in-out variant [1] with parameter $\alpha = 0.99$ to improve convergence. All lifted variants are several order-of-magnitude faster than their ground counterparts. We also find that for this particular MLN, all solutions found by the local LP formulation immediately satisfy all the cycle constraints. Closer examination reveals that this MLN prescribes attractive potentials on the pairs $(Smoke(x), Smoke(y))$, thus MAP

---

[5]The ground graphical model of this MLN has tree-width equals to the domain size.

[6]Independent result reported in [13] seems to suggest better performance can be obtained using SAUCY, a more modern tool for finding graph automorphism.

Table 1: Symmetries detection on the Friends & Smokers MLN with 10% known people. * means the process did not finish within a day.

|  |  | 10 | 20 | 50 | 100 | 200 | 1000 |
|---|---|---|---|---|---|---|---|
| Nauty | #Orbits | 12 | 23 | 25 | 27 | * | * |
|  | Time(s) | .49 | 1.77 | 172.79 | 9680.48 | * | * |
| Renaming | #Orbits | 12 | 23 | 80 | 255 | 905 | 20505 |
|  | Time(s) | .08 | .09 | .221 | .4 | .84 | 2.19 |

assignments to unknown smokers are either all true or all false.

Next, we conduct experiments with the following "Lovers & Smokers" MLN.

$$
\begin{array}{ll}
100 & Male(x) \Leftrightarrow \neg Female(x) \\
2 & Male(x) \wedge Smokes(x) \\
2 & Female(x) \wedge \neg Smokes(x) \\
0.5 & x \neq y \wedge Male(x) \wedge Female(y) \wedge Loves(x,y) \\
0.5 & x \neq y \wedge Loves(x,y) \Rightarrow (Smokes(x) \Leftrightarrow Smokes(y)) \\
-100 & x \neq y \wedge y \neq z \wedge z \neq x \wedge Loves(x,y) \wedge Loves(y,z) \wedge Loves(x,z)
\end{array}
$$

Note that this model is much more difficult because the last formula has a repulsive potential and is fully transitive. As far as we know, to date, no exact lifted inference algorithms can handle transitive clauses in polynomial time.

The first experiment assumes no evidence, a situation commonly encountered during the inference step [9] of any perceptron-style generative parameter learning method. As before, we compare local and cycle LP formulations, both ground and lifted while varying the domain size of the MLN. Figure 5(a) shows the lifted variants achieve *constant* running time regardless of the actual domain size, and are significantly more efficient than their ground counterparts as the domain size increase. Figure 5(b) illustrates how the objective value changes over cutting plane iterations (and hence time), for domain size = 5. Both the local polytope (ground and lifted) approaches have no cutting plane iterations, and hence are represented as single points. We use Integer Linear Programming (ILP) to compute a reference point of the lowest possible optimal objective value. Notice all methods are based on outer/upper bounds on the variational objective, and hence are decreasing over time. First, we can observe that the CYCLE methods converge to a solution substantially better than the LOCAL methods. However, although lifted CYCLE converges quickly, the ground CYCLE algorithm converges very slowly.

The second experiment varies the number of observed constants with random soft evidence while fixing the domain size to 100. Because ground methods do not scale to this size, we only compare lifted LOCAL and lifted CYCLE. Figure 6 shows both the running time and the obtained objective value. Observe that lifted CYCLE significantly improves the MAP objective value but at a significant computational cost when the number of observed constants increases. We note that with soft evidence, the lifted model essentially becomes a ground model which contains a large number of cycles induced by the transitive clause in the model.

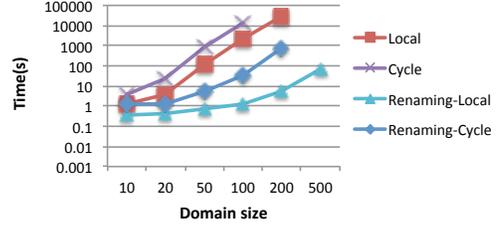

Figure 4: (Best viewed in color) "Friends & Smokers" MLN with 10% known people.

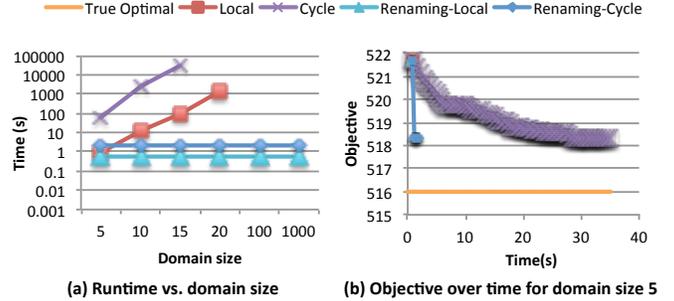

(a) Runtime vs. domain size  (b) Objective over time for domain size 5

Figure 5: (Best viewed in color) "Lovers & Smokers" MLN without evidence. The local and cycle methods did not finish within a day for larger domain sizes.

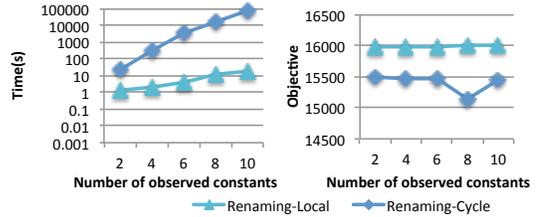

Figure 6: "Lovers & Smokers" MLN with random soft evidence, domain size = 100.

## 10 Conclusion

We presented a new general framework for lifted variational inference by introducing and studying a precise mathematical definition of symmetry of graphical models via the construction of their automorphism groups. Using the device of automorphism groups, orbits of random variables are obtained, and lifted variational inference materializes as performing the corresponding convex variational optimization problem in the space of per-orbit random variables. Our framework enables lifting a large class of approximate variational MAP inference algorithms, including the first lifted algorithm for MAP inference with cycle constraints. We presented experimental results demonstrating the clear benefits of the lifted over the ground formulations. Future extension includes how to handle approximations of the convex upper-bounds of negative entropy function $A^*$, which would enable lifting the full class of approximate convex variational marginal inference.

*Acknowledgement.* The authors gratefully acknowledge the support of the Defense Advanced Research Projects Agency (DARPA) Machine Reading Program under Air Force Research Laboratory (AFRL) prime contract no. FA8750-09-C-0181. Any opinions, findings, and conclusions or recommendations expressed in this material are those of the author(s) and do not necessarily reflect the view of DARPA, AFRL, or the U.S. government.